\documentclass[11pt]{article}

\usepackage[preprint]{acl}

\usepackage{times}
\usepackage{latexsym}
\usepackage[T1]{fontenc}
\usepackage[utf8]{inputenc}
\usepackage{microtype}
\usepackage{inconsolata}
\usepackage{graphicx}
\usepackage{booktabs}
\usepackage{stfloats}
\usepackage{amsmath}
\usepackage{amsfonts}
\usepackage{xcolor}
\usepackage{colortbl}
\usepackage{enumitem}
\usepackage{algorithm}
\usepackage{algpseudocode}

\definecolor{speedBelow}{HTML}{F4F4F2}
\definecolor{speedGainOne}{HTML}{EDF5F8}
\definecolor{speedGainLow}{HTML}{DCEBF1}
\definecolor{speedGainMid}{HTML}{C7DDE8}
\definecolor{speedGainHigh}{HTML}{AECBD9}
\definecolor{speedGainTop}{HTML}{8FB7CB}
\newcommand{\speedbelow}[1]{\cellcolor{speedBelow}#1}
\newcommand{\speedone}[1]{\cellcolor{speedGainOne}#1}
\newcommand{\speedlow}[1]{\cellcolor{speedGainLow}#1}
\newcommand{\speedmid}[1]{\cellcolor{speedGainMid}#1}
\newcommand{\speedhigh}[1]{\cellcolor{speedGainHigh}#1}
\newcommand{\speedtop}[1]{\cellcolor{speedGainTop}#1}

\title{Hybrid Verified Decoding: Learning to Allocate Verification in Speculative Decoding}

\author{
Xin Su \\
 \small{Thoughtworks} \\ \And
  Dawid Majchrowski \\
  \small{Nvidia} \\ \And
   Fangyuan Yu \\
  \small{Thoughtworks} \\ \And
    Vanshil Atul Shah \\
  \small{Nvidia} \\ \AND
     Sebastian Rogawski \\
  \small{Nvidia} \\ \And
 Pawel Morkisz \\
  \small{Nvidia} \\ \And
 Anahita Bhiwandiwalla \\
  \small{Nvidia} \\ \And
   Phillip Howard \\
  \small{Thoughtworks} \\
}

\begin{document}

\maketitle

\begin{abstract}
Large Language Model (LLM) generation remains expensive because autoregressive decoding calls the model once for each new token. Speculative decoding reduces this cost by drafting multiple tokens and verifying them with the target model in one step, but its speedup depends on how many drafted tokens are accepted. Parameter-free draft sources can propose long continuations at low cost in structured and agentic workloads, yet a cache match that looks promising at one generation step may have low payoff at the next. We propose Hybrid Verified Decoding, which predicts the accepted length of a cache draft before verification and uses this payoff estimate to choose between cache verification and a model-based drafter. Across three LLMs and sixteen datasets, Hybrid Verified Decoding is especially effective on agentic workflows, where it outperforms EAGLE3 in every setting with a 2.73x average speedup. Our analysis shows how prompt structure creates cache opportunities, how high-payoff cache drafts concentrate in a small part of the draft space, and how payoff-guided selection reduces sequential decoding work, pointing to runtime draft selection as a promising direction for speculative decoding.
\end{abstract}

\section{Introduction}

Large language model (LLM) generation is limited by autoregressive decoding: each new token requires a forward pass over the current prefix and key-value (KV) cache. Speculative decoding \citep{pmlr-v202-leviathan23a,chen2023acceleratinglargelanguagemodel} reduces this sequential cost through a draft-then-verify interface: a lower-cost source proposes candidate tokens, and the target LLM itself verifies which prefix can be committed. Its speedup depends on how many proposed tokens are accepted at each verification step.

Recent speculative decoding methods instantiate the draft source in different ways. Model-based drafters, such as the EAGLE family \citep{li2024eagle,li-etal-2024-eagle,li2025eagle3scalinginferenceacceleration}, improve draft reliability through learned proposal models. Lookup, retrieval, and cache-based methods \citep{saxena2023prompt,he-etal-2024-rest,pmlr-v235-fu24a,oliaro2025suffixdecoding} reuse continuations from the prompt, generation history, or external corpora at low proposal cost. Such reuse is common in structured and agentic workloads, where recurring context can support long cache drafts. The payoff of a cache draft, however, depends on the current generation step. A local match can produce a long draft, but a similar prefix may require different arguments, a different edit goal, or a different piece of evidence. When the target model accepts only a short prefix, the cache draft turns a cheap proposal into wasted verification.

We study this per-step decision in verified decoding and propose Hybrid Verified Decoding, which estimates the accepted length of the current cache draft before verification. If the predicted payoff is high, the cache draft is submitted for verification. Otherwise, the system uses a model-based drafter. The target model remains responsible for committed tokens, while the learned component allocates verification work across draft sources and generation steps. We train the payoff predictor from trace replay and evaluate the resulting decoder across three LLMs and sixteen datasets. Our analysis investigates the full path from input structure to cache opportunity, verification payoff, predictor selection, execution cost, and prompt-template sensitivity. Our contributions are:

\begin{itemize}[leftmargin=*, topsep=0.25em, itemsep=0.1em, parsep=0pt, partopsep=0pt]
  \item We formulate state-dependent draft selection in verified decoding as payoff-guided verification allocation, with accepted length as the payoff variable.
  \item We propose Hybrid Verified Decoding, which predicts the accepted length of the current cache draft and chooses between cache-based drafting and a model-based drafter before verification.
  \item We present a replay-based training procedure that derives payoff labels from generation traces, enabling lightweight supervision for the runtime predictor.
  \item We evaluate our method across three LLMs, sixteen datasets, and four baselines approaches spanning model-based, cache-based, rule-based hybrid, and greedy decoding.
  \item We provide a systematic analysis of cache opportunity, verification payoff, predictor behavior, execution cost, and prompt-template sensitivity, and implement a vLLM serving prototype for deployment-oriented validation.
\end{itemize}

\section{Related Work}

\paragraph{Parameter-Based Speculative Decoding.}
Parameter-based methods improve speculative decoding by adding learned proposal modules. The EAGLE family \citep{li2024eagle,li-etal-2024-eagle,li2025eagle3scalinginferenceacceleration} uses model representations to construct stronger drafts. Medusa \citep{pmlr-v235-cai24b} and PARD \citep{an2025pardacceleratingllminference} expand candidate generation through multiple heads or parallel draft modules. These methods invest additional parameters and training to improve draft reliability or candidate diversity. Hybrid Verified Decoding uses a model-based drafter as its fallback path and focuses on the verification decision made before that fallback is used.

\paragraph{Parameter-Free Speculative Decoding.}
Parameter-free methods propose draft tokens without training a separate proposal model. Prompt Lookup Decoding \citep{saxena2023prompt}, REST \citep{he-etal-2024-rest}, Lookahead Decoding \citep{pmlr-v235-fu24a}, and SuffixDecoding \citep{oliaro2025suffixdecoding} reuse continuations from the prompt, generation history, or external datastores. This line of work lowers proposal cost and fits workloads with repeated context, structured formats, or recurring intermediate states. A cache or retrieval match still leaves an open runtime question: how many of the proposed tokens will the target model accept at the current generation step? Our method builds on parameter-free drafting by estimating this accepted length before verification and using it to decide whether the current cache draft should be verified.

\paragraph{Adaptive Speculation.}
Adaptive speculation methods adjust the behavior of speculative decoding at runtime. AdaEAGLE \citep{zhang2024adaeagleoptimizingspeculativedecoding} adapts draft structures, Nightjar \citep{li2026nightjardynamicadaptivespeculative} and AdaSpec \citep{huang2026adaspecadaptivespeculativedecoding} adjust serving-time speculation, ECHO \citep{hu2026echoelasticspeculativedecoding} and TALON \citep{liu2026talonconfidenceawarespeculativedecoding} control token trees or candidate budgets, and MARS \citep{song2026marsunleashingpowerspeculative} modifies verification behavior. These methods show that speculation benefits from runtime control over draft length, tree shape, budget, or verification criteria. They leave an important gap for parameter-free drafts: after a cache match is found, the system still needs to decide whether the proposed continuation is worth verification. Hybrid Verified Decoding addresses this gap by predicting the accepted length of the proposed cache draft before verification and using that payoff estimate to choose between cache verification and a model-based drafter at each generation step.

\section{Method}

\begin{figure*}[t]
    \centering
    \includegraphics[width=0.92\textwidth]{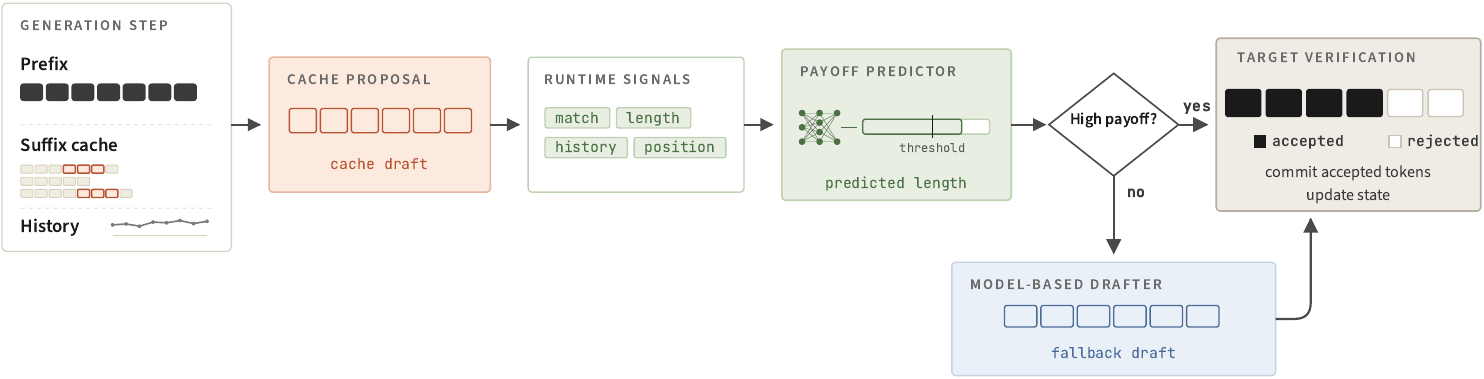}
    \caption{
    Overview of Hybrid Verified Decoding.
    }
    \label{fig:method-overview}
\end{figure*}

Speculative decoding accelerates generation when one verification step accepts multiple draft tokens. A model-based drafter improves draft reliability through training. In contrast, a parameter-free draft source reuses continuations from existing context at low proposal cost, matching the repeated structure often found in repetitive, structured, and agentic workloads. However, the quality of such drafts depends strongly on the current decoding state: an incorrect reuse may be rejected early by the target model, turning a low-cost proposal into additional verification overhead. We formulate this challenge as per-step payoff prediction: before running verification, we predict how many tokens in the current parameter-free draft are likely to be accepted by the target model, and use this prediction to decide whether to verify the draft or fall back to a model-based drafter. Figure~\ref{fig:method-overview} summarizes this workflow.

\subsection{Verified Drafting as a Common Interface}

Consider a decoding step \(t\) with current prefix \(x_{1:t}\), consisting of the input prompt and the tokens generated so far. A draft source \(D\) proposes a length-\(k\) continuation conditioned on this prefix:

\[
d_t = (\hat{x}_{t+1}, \ldots, \hat{x}_{t+k}) = D(x_{1:t}).
\]

The target model then verifies \(d_t\) and accepts a prefix of the proposal. We denote the number of accepted draft tokens by
\[
a_t(d_t) \in \{0,\ldots,k\}.
\]
Target verification also returns the token block committed after this step:
\[
(a_t(d_t), z_t) = \mathrm{Verify}(x_{1:t}, d_t; M).
\]
The first \(a_t(d_t)\) tokens of \(z_t\) match the draft prefix \((\hat{x}_{t+1}, \ldots, \hat{x}_{t+a_t(d_t)})\). If verification rejects the draft before completion, \(z_t\) also contains the target-model token at the rejection point, so decoding continues under verification.

The accepted length \(a_t(d_t)\) directly determines the benefit of a draft. A large accepted length lets one verification step commit multiple draft tokens. A small accepted length, especially when it is close to zero, leaves the system with the cost of proposing and verifying a draft but little speculative progress. Draft selection should therefore account for the expected accepted length under the current decoding state, since this quantity determines whether the cost of producing and verifying the draft can be amortized.

Our method leaves the target model and the verification rule unchanged. The only runtime choice is which candidate continuation is sent for verification. This makes draft selection a payoff estimation problem: \emph{can we predict, before verification, whether the current draft will yield a sufficiently long accepted prefix?}

\subsection{Cache-Based Drafting}

Parameter-free drafting constructs proposals by matching recurring token patterns in the input and generation history, without training a separate drafter. We instantiate this source with a suffix cache, following SuffixDecoding~\citep{oliaro2025suffixdecoding}. The cache indexes suffix matches over the input prompt and the tokens committed so far, and returns candidate continuations from matched spans. Given the current prefix \(x_{1:t}\), the cache-based drafter \(D_{\mathcal{C}}\) returns
\[
d_t^c = D_{\mathcal{C}}(x_{1:t}; \mathcal{C}),
\]
where \(\mathcal{C}\) denotes the suffix cache and \(d_t^c\) is the cache draft.

This mechanism requires no additional model forward pass. Structured and agentic workloads often contain recurring patterns across prompts and generations, creating opportunities for long cache drafts at low proposal cost. The same property also creates runtime risk: a local suffix match does not guarantee that the following continuation remains valid in the current state. For example, repeated tool-call structure may still require different arguments. If such a continuation is sent for verification, the target model may accept only a short prefix. Cache-based drafting exposes a direct tradeoff: it can produce long drafts at low proposal cost, but verification is worthwhile only when the draft is likely to yield a sufficiently long accepted prefix.

\subsection{Learning When to Verify a Cache Draft}

The value of a cache draft depends on the prefix length accepted during verification, which is unknown prior to verification. We denote it by
\[
y_t = a_t(d_t^c),
\]
where \(d_t^c\) is the cache draft at decoding step \(t\). A large \(y_t\) lets one verification step advance the generation by multiple tokens. A small \(y_t\) turns the cache proposal into verification overhead. Our goal is to estimate \(y_t\) from the current decoding state before verification.

For each cache draft, we compute a runtime feature vector
\[
\phi_t = \Phi(x_{1:t}, d_t^c, \mathcal{C}, h_t),
\]
where \(h_t\) summarizes recent decoding history. The features describe signals available before verification: the suffix match, the draft length, the current decoding position, recent cache-draft behavior, and lightweight token-structure indicators (see Appendix~\ref{app:payoff-label-predictor-training} for details). The payoff predictor maps these features to an accepted-length estimate,
\[
\hat{y}_t = g_{\theta}(\phi_t).
\]
We instantiate \(g_{\theta}\) as a lightweight multilayer perceptron over runtime features. A threshold \(\tau\) turns the estimate into a verification decision: cache drafts with \(\hat{y}_t \ge \tau\) are sent to verification, while lower-scoring states use a model-based drafter.

We derive training labels by treating each token position in a target-model generation trace as a decoding state. For a position \(t\), the visible prefix is \(x_{1:t}\). The suffix cache built from this prefix returns a cache draft \(d_t^c\). The following tokens \(x_{t+1:}\) under greedy decoding provide the target continuation for that state. The accepted-length label is the longest prefix of the cache draft that agrees with this continuation:
\[
y_t =
\max \left\{
\ell \in \{0,\ldots, |d_t^c|\} :
d_{t,1:\ell}^c = x_{t+1:t+\ell}
\right\}.
\]
We train \(g_{\theta}\) with accepted-length regression:
\[
\min_{\theta} \sum_t \left(g_{\theta}(\phi_t)-y_t\right)^2.
\]

\subsection{Hybrid Verified Decoding}
\label{sec:hybrid-verified-decoding}

For each decoding step, Hybrid Verified Decoding obtains a cache draft \(d_t^c\) and computes its predicted accepted length \(\hat{y}_t = g_{\theta}(\phi_t)\). The draft sent to verification is selected by
\[
d_t =
\begin{cases}
d_t^c, & \text{if } \hat{y}_t \ge \tau, \\
D_m(x_{1:t}), & \text{otherwise},
\end{cases}
\]
where \(D_m\) denotes the model-based drafter. Algorithm~\ref{alg:hybrid-verified-decoding} summarizes the resulting decoding loop.
The selected draft \(d_t\) is verified by the target model under the standard speculative decoding verification rule:
\[
(a_t(d_t), z_t) = \mathrm{Verify}(x_{1:t}, d_t; M),
\]
where \(M\) is the target model. The committed block \(z_t\) is appended to the output, the suffix cache is updated with \(z_t\), and decoding proceeds.

\begin{algorithm}[t]
\caption{Hybrid Verified Decoding}
\label{alg:hybrid-verified-decoding}
\begin{algorithmic}[1]
\Require target model \(M\), model-based drafter \(D_m\), suffix cache \(\mathcal{C}\), payoff predictor \(g_\theta\), threshold \(\tau\)
\State Initialize the current prefix \(x\) with the input prompt
\State Initialize \(\mathcal{C}\) from cache-warmup traces and the input prompt
\State Initialize the recent-history summary \(h\)
\While{generation is not finished}
    \State \(d^c \gets D_{\mathcal{C}}(x; \mathcal{C})\)
    \State \(\phi \gets \Phi(x, d^c, \mathcal{C}, h)\), \(\hat{y} \gets g_\theta(\phi)\)
    \If{\(\hat{y} \ge \tau\)}
        \State \(d \gets d^c\)
    \Else
        \State \(d \gets D_m(x)\)
    \EndIf
    \State \(a, z \gets \mathrm{Verify}(x, d; M)\)
    \State Append the committed block \(z\) to \(x\)
    \State Update \(\mathcal{C}\) and \(h\) with \(z\)
\EndWhile
\end{algorithmic}
\end{algorithm}

\begin{table*}[!t]
\centering
\scriptsize
\setlength{\tabcolsep}{2.2pt}
\renewcommand{\arraystretch}{1.08}
\resizebox{\textwidth}{!}{%
\begin{tabular}{llcccccccccccc}
\toprule
& & \multicolumn{4}{c}{Qwen3-8B} & \multicolumn{4}{c}{Qwen3-4B} & \multicolumn{4}{c}{Llama3.1-8B} \\
\cmidrule(lr){3-6} \cmidrule(lr){7-10} \cmidrule(lr){11-14}
Group & Dataset & G & E3 & SD & FH & G & E3 & SD & FH & G & E3 & SD & FH \\
\midrule
Agentic & Delegate-52 & \speedtop{3.25} & \speedlow{1.44} & \speedhigh{2.14} & \speedone{1.06} & \speedtop{3.40} & \speedmid{1.90} & \speedmid{1.64} & \speedone{1.18} & \speedtop{3.46} & \speedtop{3.38} & \speedbelow{0.81} & \speedlow{1.24} \\
 & InstructEdit/FineEdit & \speedtop{3.57} & \speedmid{1.51} & \speedhigh{2.30} & \speedlow{1.32} & \speedtop{4.21} & \speedhigh{2.67} & \speedmid{1.72} & \speedlow{1.22} & \speedtop{7.56} & \speedtop{5.49} & \speedone{1.15} & \speedlow{1.20} \\
\addlinespace[0.2em]
Code & RepoBench & \speedhigh{2.54} & \speedone{1.09} & \speedhigh{2.15} & \speedone{1.15} & \speedtop{3.57} & \speedmid{1.56} & \speedlow{1.44} & \speedbelow{0.81} & \speedhigh{2.12} & \speedhigh{2.99} & \speedone{1.09} & \speedlow{1.26} \\
 & SWE-bench OpenHands & \speedtop{3.70} & \speedone{1.09} & \speedtop{3.21} & \speedlow{1.28} & \speedhigh{2.50} & \speedone{1.08} & \speedmid{1.52} & \speedbelow{0.88} & \speedtop{3.31} & \speedhigh{2.11} & \speedhigh{2.60} & \speedlow{1.26} \\
 & Magicoder & \speedtop{3.31} & \speedone{1.08} & \speedtop{3.79} & \speedlow{1.34} & \speedmid{1.73} & \speedone{1.02} & \speedmid{1.82} & \speedbelow{0.87} & \speedtop{5.86} & \speedhigh{2.51} & \speedtop{4.28} & \speedlow{1.27} \\
 & MBPP & \speedtop{4.49} & \speedone{1.17} & \speedtop{4.78} & \speedmid{1.54} & \speedhigh{2.73} & \speedbelow{0.90} & \speedhigh{2.11} & \speedone{1.01} & \speedtop{3.68} & \speedbelow{0.99} & \speedtop{4.65} & \speedone{1.05} \\
\addlinespace[0.2em]
Structured & BFCL & \speedhigh{2.31} & \speedbelow{0.98} & \speedtop{3.13} & \speedlow{1.27} & \speedhigh{2.19} & \speedbelow{0.98} & \speedhigh{2.47} & \speedone{1.13} & \speedone{1.02} & \speedbelow{0.90} & \speedlow{1.32} & \speedone{1.01} \\
 & Spider & \speedtop{3.12} & \speedbelow{0.97} & \speedhigh{2.98} & \speedlow{1.33} & \speedmid{1.56} & \speedone{1.01} & \speedmid{1.79} & \speedbelow{0.88} & \speedtop{4.46} & \speedmid{1.85} & \speedtop{3.61} & \speedlow{1.24} \\
\addlinespace[0.2em]
Multi-hop QA & HotpotQA & \speedhigh{2.57} & \speedbelow{0.95} & \speedtop{3.77} & \speedlow{1.37} & \speedhigh{2.03} & \speedbelow{0.81} & \speedmid{1.97} & \speedone{1.02} & \speedhigh{2.23} & \speedbelow{0.93} & \speedhigh{2.70} & \speedbelow{0.96} \\
 & MuSiQue & \speedtop{3.55} & \speedbelow{0.95} & \speedhigh{2.88} & \speedone{1.17} & \speedhigh{2.12} & \speedone{1.18} & \speedlow{1.35} & \speedone{1.05} & \speedtop{3.00} & \speedtop{3.22} & \speedone{1.15} & \speedmid{1.97} \\
 & 2WikiMultiHopQA & \speedhigh{2.73} & \speedbelow{0.96} & \speedtop{3.28} & \speedlow{1.40} & \speedhigh{2.01} & \speedone{1.19} & \speedmid{1.65} & \speedbelow{0.98} & \speedhigh{2.52} & \speedone{1.04} & \speedhigh{2.49} & \speedbelow{0.87} \\
\addlinespace[0.2em]
Long-context & InfiniteBench & \speedtop{5.03} & \speedhigh{2.42} & \speedtop{3.04} & \speedlow{1.21} & \speedtop{5.53} & \speedhigh{2.51} & \speedhigh{2.40} & \speedone{1.16} & \speedhigh{2.23} & \speedhigh{2.84} & \speedlow{1.47} & \speedhigh{2.94} \\
 & CNN/DailyMail & \speedhigh{2.33} & \speedbelow{0.94} & \speedtop{3.85} & \speedlow{1.34} & \speedmid{1.79} & \speedbelow{0.99} & \speedhigh{2.75} & \speedone{1.02} & \speedhigh{2.43} & \speedbelow{0.94} & \speedtop{3.83} & \speedone{1.04} \\
 & GovReport & \speedhigh{2.03} & \speedbelow{0.95} & \speedtop{3.36} & \speedone{1.18} & \speedmid{1.52} & \speedone{1.01} & \speedhigh{2.08} & \speedone{1.02} & \speedbelow{0.95} & \speedlow{1.31} & \speedbelow{0.98} & \speedone{1.15} \\
\addlinespace[0.2em]
Open-ended & Alpaca & \speedhigh{2.73} & \speedbelow{0.97} & \speedtop{3.84} & \speedlow{1.40} & \speedhigh{2.06} & \speedbelow{0.85} & \speedhigh{2.40} & \speedbelow{0.83} & \speedtop{3.14} & \speedbelow{0.96} & \speedtop{5.68} & \speedone{1.05} \\
 & MT-Bench & \speedtop{3.34} & \speedbelow{0.98} & \speedtop{4.68} & \speedlow{1.41} & \speedmid{1.59} & \speedbelow{0.99} & \speedhigh{2.06} & \speedbelow{0.87} & \speedtop{3.24} & \speedone{1.18} & \speedtop{4.76} & \speedone{1.10} \\
\midrule
Avg. across datasets & & \speedtop{3.16} & \speedone{1.15} & \speedtop{3.32} & \speedlow{1.30} & \speedhigh{2.53} & \speedlow{1.29} & \speedmid{1.95} & \speedbelow{0.99} & \speedtop{3.20} & \speedhigh{2.04} & \speedhigh{2.66} & \speedlow{1.29} \\
\bottomrule
\end{tabular}%
}
\caption{Dataset-level end-to-end speedup of Hybrid Verified Decoding over four baselines across three target models. G, E3, SD, and FH denote Greedy Decoding, EAGLE3, SuffixDecoding, and Frequency-Based Hybrid Decoding. Cell shading follows speedup magnitude. The final row reports the unweighted dataset average.}
\label{tab:main-results-dataset-speedup}
\end{table*}

\section{Experiments}

\subsection{Experiment Setup}

We evaluate our method on three instruction-tuned target models: Qwen3-8B, Qwen3-4B \citep{yang2025qwen3technicalreport}, and Llama3.1-8B \citep{grattafiori2024llama3herdmodels}. Each target model is paired with its corresponding EAGLE3 checkpoint as the model-based fallback drafter. All end-to-end timing runs are measured on H100 GPUs with deterministic decoding.

Our evaluation covers 16 datasets across six workload families: agentic and workflow editing (Delegate-52, InstructEdit/FineEdit), repository-level and standalone coding (RepoBench, SWE-bench OpenHands, Magicoder, MBPP), structured generation and tool use (BFCL, Spider), multi-hop QA (HotpotQA, MuSiQue, 2WikiMultiHopQA), long-context document tasks (InfiniteBench, CNN/DailyMail, GovReport), and open-ended instruction/chat (Alpaca, MT-Bench). These workloads differ in input length, structural constraints, context reuse, and output format.

All methods are evaluated inside the generation loop of the official EAGLE repository\footnote{\url{https://github.com/SafeAILab/EAGLE}} and use the chat template associated with each target model. We report run-level throughput as generated tokens per second, and define speedup as the throughput ratio relative to the corresponding baseline.

\subsection{Baselines}
\label{sec:baselines}

We compare our hybrid verified decoding method from Section~\ref{sec:hybrid-verified-decoding} with four baselines: target-only decoding, model-based speculative decoding, cache-based drafting, and frequency-based hybrid decoding.

\paragraph{Greedy Decoding.}
Greedy decoding is the target-only reference. It uses neither an EAGLE3 drafter nor a suffix cache, and provides the wall-clock throughput without speculative decoding.

\paragraph{EAGLE3.}
EAGLE3~\citep{li2025eagle3scalinginferenceacceleration} is a strong model-based speculative decoding baseline. It uses the EAGLE3 checkpoint corresponding to the target model to produce drafts, which are then verified by the same target model.

\paragraph{SuffixDecoding.}
SuffixDecoding~\citep{oliaro2025suffixdecoding} is the cache-based drafting baseline. It proposes drafts from a suffix cache and submits them for verification.

\paragraph{Frequency-Based Hybrid Decoding.}
Frequency-Based Hybrid Decoding is a rule-based hybrid baseline. It uses the same cache-first, EAGLE3-fallback structure as our method, with the verification decision based on the frequency-derived cache score from SuffixDecoding~\citep{oliaro2025suffixdecoding}.

\subsection{Main Results}

Table~\ref{tab:main-results-dataset-speedup} reports end-to-end speedup for Hybrid Verified Decoding across 16 datasets and three target models. Across the 48 model-dataset pairs, our method is faster than Greedy Decoding in 47 cases and faster than SuffixDecoding in 46 cases, with average speedups of 2.97x and 2.64x, respectively. These comparisons support the central design motivation: cache drafts can substantially reduce sequential decoding steps when accepted by the target model, while low-payoff cache drafts should be filtered before they introduce additional verification cost.

Agentic and workflow-style workloads make this trade-off especially concrete. On Delegate-52 and InstructEdit/FineEdit, Hybrid Verified Decoding outperforms both EAGLE3 and Frequency-Based Hybrid Decoding for all three target models. These workloads repeatedly expose tool-call structure, editing context, repository state, and intermediate outputs, creating opportunities for long cache drafts. At the same time, each step remains state-dependent: a similar tool call or editing context may require a different argument or a different target change. The learned payoff predictor addresses this state-dependent decision by estimating, before verification, whether the current cache draft is likely to yield a sufficiently long accepted prefix.

The model-level differences follow the cost trade-off identified in recent analyses of speculative decoding~\citep{liu2026speculativeperformanceillusion,yan-etal-2025-decoding}: speedup depends jointly on accepted length, drafting cost, verification cost, and system overhead, while acceptance behavior varies across requests, token positions, and datasets. On Qwen3-4B, Hybrid Verified Decoding still improves consistently over Greedy Decoding and SuffixDecoding, but its margin over Frequency-Based Hybrid Decoding is smaller than on Qwen3-8B and Llama3.1-8B. Smaller target models reduce the relative cost of verification, so the benefit of a learned switching policy becomes more visible when it clearly increases accepted length or avoids wasted cache verification. The results show that Hybrid Verified Decoding is strongest when cache opportunities appear frequently, yet draft correctness still depends on the current decoding state. Section~\ref{sec:analysis} examines this pattern from input structure to cache opportunity, payoff prediction, execution cost, hardware behavior, serving integration, and prompt-template sensitivity.

\section{Analysis}
\label{sec:analysis}

\subsection{What Makes a Task Cache-Friendly?}

\begin{figure*}[!t]
    \centering
    \begin{minipage}[t]{0.58\textwidth}
        \centering
        \includegraphics[width=\linewidth]{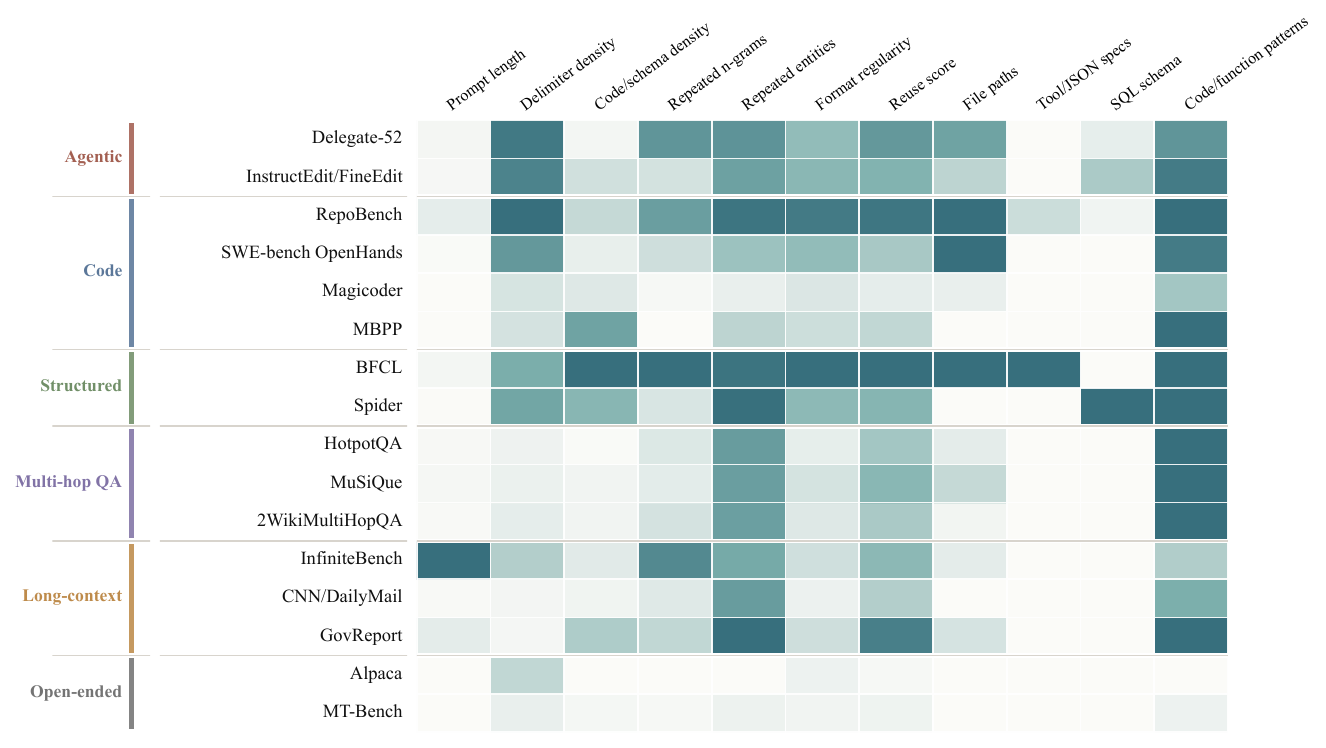}
        \vspace{-0.6em}
        {\small (a) Prompt-side structure.}
    \end{minipage}\hfill
    \begin{minipage}[t]{0.38\textwidth}
        \centering
        \includegraphics[width=\linewidth]{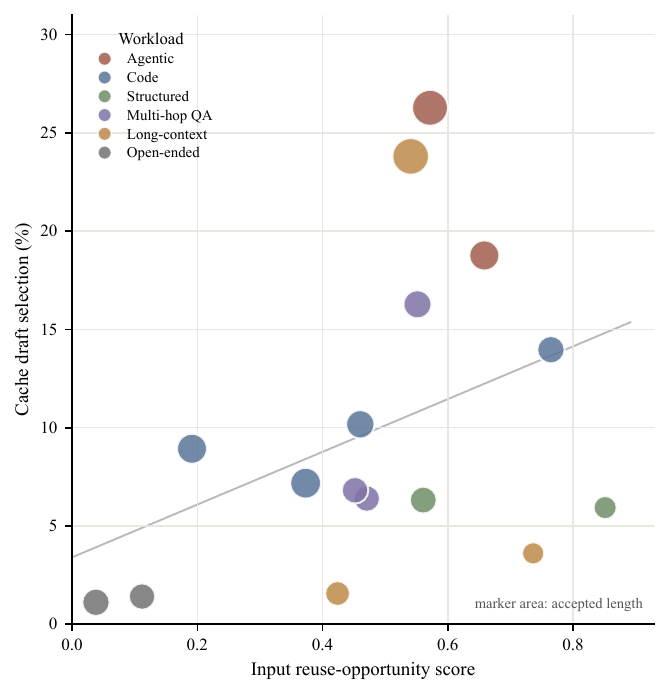}
        \vspace{-0.6em}
        {\small (b) Cache-draft selection.}
    \end{minipage}
    \caption{
    Prompt-side cache opportunity and runtime cache-draft behavior.
    Panel (a) reports normalized input structure features by workload family. Darker cells indicate larger feature values after column-wise normalization.
    Panel (b) reports cache-draft selection rate against prompt-side reuse opportunity, with marker area scaled by mean accepted length.
    }
    \label{fig:analysis-cache-opportunity}
\end{figure*}

The prompt-side view in Figure~\ref{fig:analysis-cache-opportunity}(a) shows that cache opportunity varies substantially across workload families. Agentic/workflow editing and code/repository workloads expose dense structural signals, including repeated entities, file paths, delimiters, and code/function patterns. Structured generation workloads show stronger schema-like and tool-oriented formats, while long-context workloads provide extended context and recurring evidence. Open-ended chat prompts are sparser along these structural dimensions. These patterns suggest that cache-based drafting benefits from inputs that repeatedly expose format, entities, contextual constraints, or workflow state, since these structures provide matchable cues for future continuations.

The runtime cache-draft view in Figure~\ref{fig:analysis-cache-opportunity}(b) shows a similar trend. Workloads with stronger prompt-side reuse structure tend to trigger cache drafts more often, with agentic/workflow editing and long-context tasks occupying the higher-selection region and open-ended chat tasks remaining near the lower-selection region. This correspondence indicates that cache opportunity comes from recurring structure, stable format, context recurrence, and workflow state. Prompt-side structure explains where cache drafts originate. The accepted length of a selected draft still depends on whether the continuation matches the current decoding state. We therefore next examine how cache opportunity turns into verification payoff.

\subsection{When Do Cache Drafts Become Worth Verifying?}

\begin{figure}[!t]
    \centering
    \includegraphics[width=\linewidth]{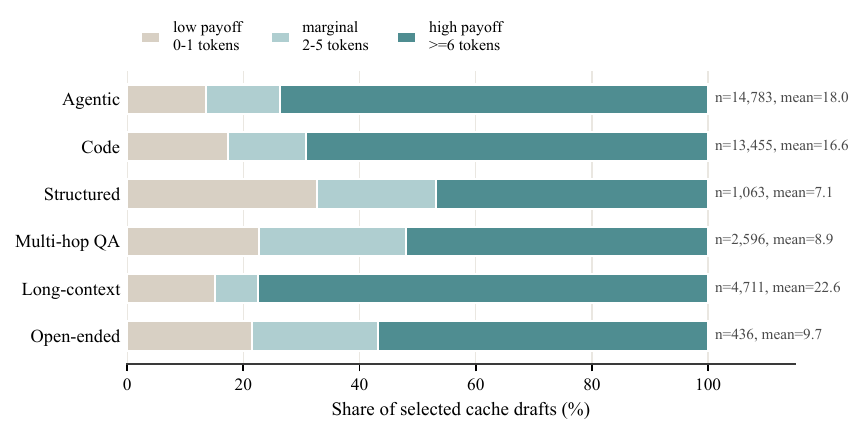}
    \caption{
    Verification payoff distribution of selected cache drafts by workload family.
    Bars group drafts by realized accepted prefix length, and annotations report selected draft count and mean accepted length.
    }
    \label{fig:analysis-cache-payoff-distribution}
\end{figure}

Figure~\ref{fig:analysis-cache-payoff-distribution} shows the verification payoff after a cache draft has been selected. The payoff distribution differs sharply across workload families. Agentic, code, and long-context workloads contain a larger share of high-payoff cache drafts, and their selected cache drafts reach longer mean accepted lengths. In these workloads, cache matches more often extend into a long accepted prefix. Structured and multi-hop QA workloads contain a larger fraction of low-payoff and marginal-payoff drafts. Open-ended workloads still contain some high-payoff cache drafts, but they produce far fewer selected cache drafts overall.

This distribution separates cache opportunity from verification payoff. A cache match can provide a candidate continuation, but its value is determined by how many of its tokens the target model accepts in the current decoding state. High-payoff drafts amortize verification cost over multiple committed tokens. Low-payoff drafts add verification work while advancing generation by only a short prefix. Hybrid Verified Decoding therefore estimates the payoff of the current cache draft before verification and reserves cache verification for drafts with sufficient expected payoff. The next section analyzes whether the payoff predictor identifies these high-payoff drafts.

\subsection{Can the Predictor Identify High-Payoff Drafts?}
\label{sec:predictor-payoff-selection}

We enumerate the proposed cache drafts returned by the suffix cache on test prompts, and compute their realized accepted length \(y_t\) from the target-model trace. Drafts with \(y_t \ge 6\) are treated as oracle high-payoff drafts. This oracle label is used for analysis and is not available during runtime decoding. The payoff predictor scores the same drafts with \(\hat{y}_t\), and drafts with \(\hat{y}_t \ge 6\) form the predictor-selected set. We then use oracle high-payoff drafts as the positive set and report the precision and recall of predictor selection. Appendix~\ref{app:analysis-metric-definitions} defines the corresponding metrics.

\begin{table}[!t]
\centering
\small
\setlength{\tabcolsep}{3.2pt}
\renewcommand{\arraystretch}{1.08}
\resizebox{\linewidth}{!}{%
\begin{tabular}{lrrrr}
\toprule
Model & \(P(y_t \ge 6)\) & \(P(\hat{y}_t \ge 6)\) & Precision & Recall \\
\midrule
Qwen3-8B & 4.8 & 3.8 & 78.2 & 61.4 \\
Qwen3-4B & 6.6 & 5.4 & 82.9 & 67.8 \\
Llama3.1-8B & 8.9 & 8.1 & 87.6 & 79.8 \\
\bottomrule
\end{tabular}
}
\caption{
Payoff-prediction quality on proposed cache drafts from test prompts.
All values are percentages.
Oracle high-payoff drafts satisfy \(y_t \ge 6\), and predictor-selected drafts satisfy \(\hat{y}_t \ge 6\).
}
\label{tab:predictor-payoff-selection}
\end{table}

Table~\ref{tab:predictor-payoff-selection} shows that cache-draft payoff is concentrated in a small portion of decoding states. Across the three target models, 4.8\%, 6.6\%, and 8.9\% of proposed cache drafts meet the oracle high-payoff criterion. The predictor selects a similarly sized set, covering 3.8\%, 5.4\%, and 8.1\% of proposed drafts. This selected set closely matches the oracle high-payoff set: precision reaches 78.2\%, 82.9\%, and 87.6\%, while recall reaches 61.4\%, 67.8\%, and 79.8\%.

The analysis reframes the cache decision as state selection. Cache drafts create many verification opportunities, but the payoff is concentrated in states where the continuation aligns with the current generation context. The payoff predictor identifies much of this high-payoff region before verification, allowing the hybrid decoder to focus cache verification on drafts that are more likely to amortize verification cost.

\subsection{How Do Draft Decisions Translate into Speedup?}

\begin{figure}[!t]
    \centering
    \includegraphics[width=\linewidth]{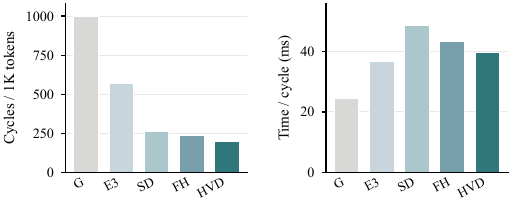}
    \caption{
    Execution breakdown on Llama3.1-8B.
    HVD denotes Hybrid Verified Decoding.
    Left: decoding cycles per 1K output tokens.
    Right: wall-clock time per decoding cycle.
    Lower values are better in both panels.
    }
    \label{fig:analysis-execution-breakdown}
\end{figure}

\paragraph{Execution Breakdown.}
Figure~\ref{fig:analysis-execution-breakdown} breaks down the execution path on Llama3.1-8B. Hybrid Verified Decoding reduces the number of decoding cycles needed per 1K output tokens from 571 for EAGLE3, 264 for SuffixDecoding, and 235 for Frequency-Based Hybrid Decoding to 200. The learned rule converts cache-draft selection into fewer sequential verification steps.

The lower cycle count is accompanied by controlled per-cycle cost. Our method runs at 39.6 ms per cycle, lower than SuffixDecoding (48.4 ms) and Frequency-Based Hybrid Decoding (43.3 ms), and close to EAGLE3 (36.7 ms). The speedup comes from reducing sequential decoding work while keeping verification-step cost controlled.

This breakdown connects draft selection to system-level speed. Hybrid Verified Decoding reduces the number of verification cycles while keeping the cost of each cycle close to the model-based drafter and below the cache-only and rule-based paths. The throughput gain comes from an execution shift: verification work is spent on decoding states that commit more tokens per cycle.

\paragraph{Cross-GPU Behavior.}
\begin{figure}[!t]
    \centering
    \includegraphics[width=\linewidth]{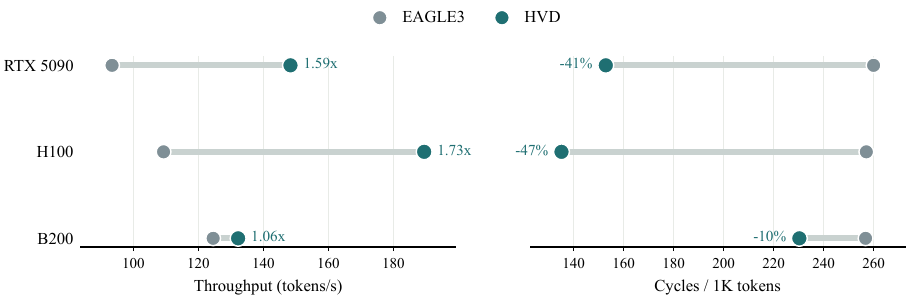}
    \caption{
    Cross-GPU case study on Qwen3-8B InfiniteBench.
    The experiment uses the same four prompts, prompt token limit, and decoding budget on RTX 5090, H100, and B200.
    Left: end-to-end throughput.
    Right: decoding cycles per 1K generated tokens.
    }
    \label{fig:analysis-hardware-case}
\end{figure}

We use a small cross-GPU case study to compare EAGLE3 and Hybrid Verified Decoding under the same setting: Qwen3-8B on InfiniteBench with the same four prompts and decoding budget. Figure~\ref{fig:analysis-hardware-case} reports both end-to-end throughput and cycles per 1K generated tokens. Relative to EAGLE3, Hybrid Verified Decoding reaches 1.59x and 1.73x speedup on RTX 5090 and H100, where cycles per 1K tokens drop from 260.0 to 152.9 and from 257.1 to 135.2, respectively. On B200, the speedup is 1.06x, with cycles per 1K tokens decreasing from 256.8 to 230.3. These results suggest that Hybrid Verified Decoding is particularly effective on consumer and standard server GPUs, where reducing verification cycles can translate into substantial throughput gains.

\paragraph{vLLM Prototype.}
\begin{table}[!t]
\centering
\small
\setlength{\tabcolsep}{6pt}
\renewcommand{\arraystretch}{1.08}
\begin{tabular}{lrr}
\toprule
Dataset & vs. EAGLE3 & vs. SuffixDecoding \\
\midrule
MT-Bench & 2.31x & 1.21x \\
CNN/DailyMail & 3.56x & 1.16x \\
\bottomrule
\end{tabular}
\caption{vLLM serving case study on Qwen3-8B.}
\label{tab:vllm-serving-case}
\end{table}

We further validate Hybrid Verified Decoding in the vLLM \citep{kwon2023efficient} serving stack. The serving setting tests whether payoff-guided switching remains effective under serving-time scheduling and execution. As shown in Table~\ref{tab:vllm-serving-case}, Hybrid Verified Decoding is 2.31x and 3.56x faster than EAGLE3, and 1.21x and 1.16x faster than SuffixDecoding, on MT-Bench and CNN/DailyMail respectively. These results show that payoff-guided switching extends beyond the offline generation loop and delivers clear throughput gains in a mainstream serving system. We provide implementation details for the vLLM prototype in Appendix~\ref{app:vllm-serving-prototype}.

\subsection{How Does Prompt Construction Shape Draft Payoff?}

We run a controlled prompt-template diagnostic on Llama3.1-8B MT-Bench, changing only the prompt construction while keeping the rest of the setup fixed.
\begin{table}[!t]
\centering
\small
\setlength{\tabcolsep}{4pt}
\renewcommand{\arraystretch}{1.08}
\resizebox{\linewidth}{!}{%
\begin{tabular}{llrrr}
\toprule
Prompt construction & Method & Mean accepted & Cycles / 1K & Throughput \\
 & & tokens \(\uparrow\) & tokens \(\downarrow\) & \(\uparrow\) \\
\midrule
Template omitted & EAGLE3 & 1.20 & 455 & 55.6 \\
Official template & EAGLE3 & 4.21 & 192 & 134.4 \\
Template omitted & Hybrid Verified Decoding & 7.32 & 120 & 194.2 \\
\bottomrule
\end{tabular}
}
\caption{
Prompt-template diagnostic on Llama3.1-8B MT-Bench.
Template omitted uses the same request content without the Llama chat template.
Official template applies the Llama chat template before tokenization.
Throughput is tokens per second.
}
\label{tab:prompt-template-diagnostic}
\end{table}
Table~\ref{tab:prompt-template-diagnostic} isolates prompt-template sensitivity. When the Llama chat template is omitted, EAGLE3 reaches only 1.20 mean accepted tokens and requires 455 cycles per 1K tokens. Applying the official template raises mean accepted tokens to 4.21 and reduces cycles to 192, showing how strongly a learned model-based drafter can depend on prompt construction. Under the same template-omitted setting, Hybrid Verified Decoding reaches 7.32 accepted tokens and 120 cycles per 1K tokens. This result highlights the value of runtime draft selection: when the model-based drafter loses payoff under a prompt-construction mismatch, cache drafts can still provide high-payoff verification paths.

\section{Conclusion}

In this paper, we introduce Hybrid Verified Decoding, a payoff-guided method for state-dependent draft selection in speculative decoding. It predicts the accepted length of a cache draft before verification and uses this estimate to choose between cache verification and a model-based drafter. Across target models and task families, our experiments show improved end-to-end throughput and concentrated cache verification on high-payoff decoding states. These results point to a broader direction for speculative decoding, where runtime systems decide which drafts merit verification while continuing to build stronger drafters.

\section*{Limitations}
\label{sec:limitations}

Our experiments cover two model families, Qwen and Llama, and instantiate hybrid selection as a choice between suffix-cache drafting and EAGLE3 model-based drafting. We have not systematically evaluated additional model families or broader draft-source combinations, such as multi-head drafters.

\bibliography{anthology,custom}

\appendix

\section{Experimental Details}
\label{app:experimental-details}

All end-to-end evaluations use the same generation loop, prompt tokenization, target model, decoding budget, and hardware class for a given model-dataset setting.

\paragraph{Models.}
Table~\ref{tab:app-model-checkpoints} lists the Hugging Face\footnote{\url{https://huggingface.co/}} checkpoints used for target models and EAGLE3 drafters.

\begin{table}[h]
\centering
\small
\setlength{\tabcolsep}{3pt}
\renewcommand{\arraystretch}{1.08}
\resizebox{\linewidth}{!}{%
\begin{tabular}{lll}
\toprule
Model & Target model checkpoint & EAGLE3 checkpoint \\
\midrule
Qwen3-8B & Qwen/Qwen3-8B & Tengyunw/qwen3\_8b\_eagle3 \\
Qwen3-4B & Qwen/Qwen3-4B & AngelSlim/Qwen3-4B\_eagle3 \\
Llama3.1-8B & meta-llama/Llama-3.1-8B-Instruct & yuhuili/EAGLE3-LLaMA3.1-Instruct-8B \\
\bottomrule
\end{tabular}
}
\caption{Target model and EAGLE3 checkpoints.}
\label{tab:app-model-checkpoints}
\end{table}

\paragraph{Generation Configuration.}
Table~\ref{tab:app-generation-config} lists the shared generation configuration.

\begin{table}[h]
\centering
\small
\setlength{\tabcolsep}{5pt}
\renewcommand{\arraystretch}{1.08}
\begin{tabular}{lr}
\toprule
Setting & Value \\
\midrule
Batch size & 1 \\
Maximum new tokens & 8192 \\
Maximum sequence length & 16384 \\
Temperature & 0.0 \\
Data type & float16 \\
Split seed & 42 \\
\bottomrule
\end{tabular}
\caption{Shared generation configuration.}
\label{tab:app-generation-config}
\end{table}

\paragraph{Drafting Configuration.}
Table~\ref{tab:app-drafting-config} lists the EAGLE3 and suffix-cache configuration.

\begin{table}[h]
\centering
\small
\setlength{\tabcolsep}{4pt}
\renewcommand{\arraystretch}{1.08}
\begin{tabular}{llr}
\toprule
Component & Setting & Value \\
\midrule
EAGLE3 & Total draft tokens & 32 \\
EAGLE3 & Tree depth & 8 \\
EAGLE3 & Top-k & 4 \\
Suffix cache & Maximum tree depth & 64 \\
Suffix cache & Maximum draft tokens & 64 \\
Suffix cache & Maximum speculative factor & 1.0 \\
Suffix cache & Maximum speculative offset & 0.0 \\
Suffix cache & Minimum token probability & 0.1 \\
\bottomrule
\end{tabular}
\caption{Drafting configuration.}
\label{tab:app-drafting-config}
\end{table}

\paragraph{Hybrid Configuration.}
Frequency-Based Hybrid Decoding uses the default suffix-cache score of the SuffixDecoding baseline described in Section~\ref{sec:baselines}. Hybrid Verified Decoding uses a lightweight payoff predictor over runtime features available before verification. The cache draft is selected when the predicted accepted length is at least 6. Predictor features, architecture, and training details are given in Appendix~\ref{app:payoff-label-predictor-training}.

\paragraph{Evaluation Metrics.}
Main end-to-end results are measured on H100 GPUs. Including trace preparation, payoff-predictor training, and the reported timing, diagnostic, and hardware runs, the experiments used approximately 50 GPU-hours. For each run, we record generated token count \(N\), wall-clock decoding time \(T\), and decoding-cycle count \(C\). Throughput is computed as
\[
\mathrm{Throughput} = \frac{N}{T}.
\]
For a method \(m\) and baseline \(b\) evaluated on the same model, dataset, and prompt set, speedup is computed as
\[
\mathrm{Speedup}(m,b) = \frac{\mathrm{Throughput}_m}{\mathrm{Throughput}_b}.
\]
For execution analysis, we also report decoding cycles per 1K generated tokens,
\[
\mathrm{CyclesPer1K} = 1000 \cdot \frac{C}{N},
\]
and average time per decoding cycle,
\[
\mathrm{TimePerCycle} = 1000 \cdot \frac{T}{C},
\]
where \(\mathrm{TimePerCycle}\) is measured in milliseconds. The aggregate metrics reported in the paper are computed from the recorded run outputs.

\begin{table*}[t]
\centering
\small
\setlength{\tabcolsep}{5pt}
\renewcommand{\arraystretch}{1.08}
\begin{tabular}{p{0.20\textwidth}p{0.24\textwidth}p{0.45\textwidth}}
\toprule
Workload family & Dataset & Source \\
\midrule
Agentic / workflow editing & Delegate-52 & \texttt{microsoft/delegate52} \\
 & InstructEdit/FineEdit & \texttt{YimingZeng/FineEdit\_bench} \\
\midrule
Code / repository & RepoBench & \texttt{tianyang/repobench\_python\_v1.1} \\
 & SWE-bench OpenHands & \texttt{SWE-bench/SWE-bench\_Verified} \\
 & Magicoder & \texttt{ise-uiuc/Magicoder-Evol-Instruct-110K} \\
 & MBPP & \texttt{google-research-datasets/mbpp} \\
\midrule
Structured generation & BFCL & \texttt{llamastack/bfcl\_v3} \\
 & Spider & \texttt{NBAmine/xlangai-spider-with-context} \\
\midrule
Multi-hop QA & HotpotQA & \texttt{hotpotqa/hotpot\_qa} \\
 & MuSiQue & \texttt{dgslibisey/MuSiQue} \\
 & 2WikiMultiHopQA & \texttt{Alab-NII/2wikimultihop} \\
\midrule
Long-context & InfiniteBench & \texttt{xinrongzhang2022/InfiniteBench} \\
 & CNN/DailyMail & \texttt{abisee/cnn\_dailymail} \\
 & GovReport & \texttt{ccdv/govreport-summarization} \\
\midrule
Open-ended instruction / chat & Alpaca & \texttt{tatsu-lab/alpaca} \\
 & MT-Bench & Official EAGLE GitHub repository (MT-Bench prompts) \\
\bottomrule
\end{tabular}
\caption{Dataset sources used to construct the evaluation prompt package.}
\label{tab:app-dataset-sources}
\end{table*}

\section{Payoff Label Construction and Predictor Training}
\label{app:payoff-label-predictor-training}

\paragraph{Payoff Label Construction.}
We construct training examples by replaying cache lookup on recorded target-model generations. For each trace, we treat each token position \(t\) as a decoding state with prefix \(x_{1:t}\). The suffix cache is queried with this prefix and returns a cache draft \(d_t^c\). The label is the number of draft tokens that match the following tokens in the same trace:
\[
y_t =
\max \left\{
\ell \in \{0,\ldots, |d_t^c|\} :
d_{t,1:\ell}^c = x_{t+1:t+\ell}
\right\}.
\]
The recorded continuation therefore provides the accepted-length label for the cache draft, without additional target-model inference for each replayed draft.

\paragraph{Predictor Features.}
The predictor uses only signals available before verification. These features describe the cache match that produced the draft, the current decoding state, recent cache-draft behavior, and simple token patterns in the proposed draft. Table~\ref{tab:app-predictor-features} groups these features into cache-match, decode-state, recent-history, and draft-structure signals. The feature set is designed to capture whether the current draft is likely to remain valid in the current decoding state, without using the target model's verification result for that draft.

\begin{table}[h]
\centering
\small
\setlength{\tabcolsep}{5pt}
\renewcommand{\arraystretch}{1.12}
\begin{tabular}{p{0.27\linewidth}p{0.64\linewidth}}
\toprule
Feature group & Description \\
\midrule
Cache match & Suffix score, matched length, draft length \\
Decode state & Prompt length, generated length, relative decode position \\
Recent history & Recent cache-draft selection rate and accepted lengths over recent decoding steps \\
Draft structure & Indicators for whitespace, punctuation, line breaks, brackets, and delimiter-like tokens \\
\bottomrule
\end{tabular}
\caption{Payoff predictor feature groups.}
\label{tab:app-predictor-features}
\end{table}

\paragraph{Predictor Training.}
The predictor maps the pre-verification feature vector \(\phi_t\) to a predicted accepted length:
\[
\hat{y}_t = g_\theta(\phi_t).
\]
We train it with squared-error regression:
\[
\min_\theta \sum_t \left(g_\theta(\phi_t) - y_t\right)^2.
\]
The predictor is a small multilayer perceptron. Table~\ref{tab:app-predictor-training-config} lists the training configuration.

\begin{table}[h]
\centering
\small
\setlength{\tabcolsep}{5pt}
\renewcommand{\arraystretch}{1.08}
\begin{tabular}{lr}
\toprule
Setting & Value \\
\midrule
Hidden layers & 2 \\
Hidden size & 256 \\
Activation & ReLU \\
Learning rate & 0.001 \\
Batch size & 65536 \\
Epochs & 20 \\
Seed & 42 \\
\bottomrule
\end{tabular}
\caption{Payoff predictor training configuration.}
\label{tab:app-predictor-training-config}
\end{table}

\paragraph{Runtime Use.}
At each decoding step, the same pre-verification features are computed for the current cache draft. Hybrid Verified Decoding selects the cache draft when the predicted accepted length is at least 6. Otherwise, it uses the model-based drafter. The target model and verification rule are unchanged.

\section{Additional End-to-End Results}
\label{app:additional-end-to-end-results}

Tables~\ref{tab:app-end2end-runtime-qwen8}--\ref{tab:app-end2end-runtime-llama} report aggregate measurements over the sixteen datasets.

\begin{table}[H]
\centering
\scriptsize
\setlength{\tabcolsep}{3pt}
\renewcommand{\arraystretch}{1.05}
\begin{tabular}{lrrrrr}
\toprule
Method & Throughput & Cycles & Accepted & Cache & EAGLE3 \\
 & tokens/s & /1K & tokens & \% & \% \\
\midrule
Greedy & 39.9 & 998.7 & -- & -- & -- \\
EAGLE3 & 109.5 & 240.2 & 3.16 & 0.0 & 100.0 \\
Suffix & 42.1 & 397.8 & 1.51 & 97.6 & 0.0 \\
Frequency Hybrid & 105.0 & 188.4 & 4.31 & 8.0 & 92.0 \\
Hybrid Verified & 133.0 & 189.7 & 4.27 & 9.7 & 90.3 \\
\bottomrule
\end{tabular}
\caption{Aggregate results for Qwen3-8B. Columns report throughput, cycles per 1K output tokens, mean accepted length, and draft-source fractions. Dashes mark fields outside the greedy path.}
\label{tab:app-end2end-runtime-qwen8}
\end{table}

\begin{table}[H]
\centering
\scriptsize
\setlength{\tabcolsep}{3pt}
\renewcommand{\arraystretch}{1.05}
\begin{tabular}{lrrrrr}
\toprule
Method & Throughput & Cycles & Accepted & Cache & EAGLE3 \\
 & tokens/s & /1K & tokens & \% & \% \\
\midrule
Greedy & 38.3 & 1000.0 & -- & -- & -- \\
EAGLE3 & 76.7 & 309.7 & 2.23 & 0.0 & 100.0 \\
Suffix & 54.6 & 346.1 & 1.89 & 97.6 & 0.0 \\
Frequency Hybrid & 104.5 & 223.9 & 3.47 & 8.2 & 91.8 \\
Hybrid Verified & 101.1 & 228.6 & 3.37 & 8.8 & 91.2 \\
\bottomrule
\end{tabular}
\caption{Aggregate results for Qwen3-4B, using the column definitions in Table~\ref{tab:app-end2end-runtime-qwen8}.}
\label{tab:app-end2end-runtime-qwen4}
\end{table}

\begin{table}[H]
\centering
\scriptsize
\setlength{\tabcolsep}{3pt}
\renewcommand{\arraystretch}{1.05}
\begin{tabular}{lrrrrr}
\toprule
Method & Throughput & Cycles & Accepted & Cache & EAGLE3 \\
 & tokens/s & /1K & tokens & \% & \% \\
\midrule
Greedy & 40.7 & 1000.0 & -- & -- & -- \\
EAGLE3 & 47.7 & 571.0 & 0.75 & 0.0 & 100.0 \\
Suffix & 78.2 & 264.0 & 2.79 & 93.9 & 0.0 \\
Frequency Hybrid & 98.3 & 234.9 & 3.26 & 9.4 & 90.6 \\
Hybrid Verified & 126.2 & 200.1 & 4.00 & 18.6 & 81.4 \\
\bottomrule
\end{tabular}
\caption{Aggregate results for Llama3.1-8B, using the column definitions in Table~\ref{tab:app-end2end-runtime-qwen8}.}
\label{tab:app-end2end-runtime-llama}
\end{table}

\section{Dataset and Prompt Details}
\label{app:dataset-prompt-details}

We evaluate on sixteen datasets grouped into six workload families. For every dataset, we fix 20 prompts for reported timing and analysis, giving 320 measured prompts per target model. Separate cache-warmup prompts from the same dataset are used to initialize the suffix cache and construct replay-based payoff labels. The test prompts are never used for cache warmup or predictor training.

For Hugging Face datasets, Table~\ref{tab:app-dataset-sources} reports the dataset identifier. For other sources, it reports the project or benchmark source used to construct the prompt package.

\section{vLLM Serving Prototype}
\label{app:vllm-serving-prototype}

We implement Hybrid Verified Decoding in the vLLM speculative decoding path. For each request, the serving path runs the selected draft source, verifies the proposed tokens with the target model, and updates the decoding state used by the payoff predictor. The predictor scores cache drafts before verification and selects cache verification when the predicted accepted length reaches the threshold. Otherwise, the request uses the model-based drafter.

The serving case study uses Qwen3-8B with its EAGLE3 checkpoint and the same payoff predictor used in the offline experiments. We run batch size 1 serving on H100 and evaluate MT-Bench and CNN/DailyMail under deterministic decoding. Table~\ref{tab:app-vllm-serving-config} summarizes the serving configuration.

\begin{table}[h]
\centering
\small
\setlength{\tabcolsep}{5pt}
\renewcommand{\arraystretch}{1.08}
\begin{tabular}{ll}
\toprule
Setting & Value \\
\midrule
Serving engine & vLLM \\
Target model & Qwen3-8B \\
Model-based drafter & EAGLE3 \\
Cache drafter & Suffix cache \\
Predictor decision & Predicted accepted length \(\ge 6\) \\
Batch size & 1 \\
Hardware & H100 \\
Datasets & MT-Bench, CNN/DailyMail \\
\bottomrule
\end{tabular}
\caption{Configuration for the vLLM serving case study.}
\label{tab:app-vllm-serving-config}
\end{table}

\section{Artifact Licenses and Intended Use}
\label{app:artifact-licenses}

We use publicly available model checkpoints, datasets, benchmarks, and code artifacts for research evaluation of decoding efficiency under their respective licenses and access terms. The Qwen checkpoints are released under Apache-2.0, and Llama3.1-8B-Instruct uses the Llama 3.1 Community License. The EAGLE repository and vLLM are released under Apache-2.0. Dataset licenses and source terms vary across benchmarks, including Apache-2.0, MIT, CC-BY, CC-BY-SA, CC-BY-NC, and CDLA-Permissive terms. Our code will be released under the MIT License.

\section{Analysis Metric Definitions}
\label{app:analysis-metric-definitions}

In Section~\ref{sec:predictor-payoff-selection} and Table~\ref{tab:predictor-payoff-selection}, a proposed cache draft is an oracle high-payoff draft when its replayed accepted length is at least 6. A predictor-selected draft has predicted accepted length at least 6. Precision is the fraction of predictor-selected drafts that are oracle high-payoff drafts, and recall is the fraction of oracle high-payoff drafts selected by the predictor. Cache selection rate is the fraction of decoding cycles in which Hybrid Verified Decoding submits a cache draft for verification. Mean accepted length is averaged over verified draft cycles for the corresponding method or draft source.

\section{AI Assistant Use}
\label{app:ai-assistant-use}

AI assistants were used to support coding and text editing. Their use followed the ACL policy on publication ethics. The authors reviewed the generated suggestions and are responsible for the submitted work.

\end{document}